\pdfoutput=1

\documentclass[11pt]{article}

\usepackage[preprint]{acl}
\usepackage[normalem]{ulem}
\useunder{\uline}{\ul}{}
\usepackage{times}
\usepackage{latexsym}
 \usepackage{multirow}
\usepackage[T1]{fontenc}

\usepackage[utf8]{inputenc}

\usepackage{microtype}
\usepackage{amssymb}
\usepackage{inconsolata}
\usepackage[normalem]{ulem}
\useunder{\uline}{\ul}{}
\usepackage{algorithm}
\usepackage{algpseudocode}
\usepackage{booktabs}
\usepackage{multirow}
\usepackage{amsmath} 
\usepackage{enumitem}
\usepackage{color,xcolor}
\usepackage{tabularx}
\usepackage[most]{tcolorbox}
\definecolor{light-blue}{RGB}{135,206,250}
\definecolor{light-pink}{RGB}{255,192,203}
\usepackage{caption}
\usepackage{graphicx}
\tcbuselibrary{breakable}
\title{Knowledge Editing through Chain-of-Thought}
\usepackage{authblk}
\author[1,2]{\textbf{Changyue Wang}\thanks{cy-wang24@mails.tsinghua.edu.cn}}
\author[1]{\textbf{Weihang Su}}
\author[2,1]{\textbf{Qingyao Ai}\thanks{Corresponding Author: aiqy@tsinghua.edu.cn}}
\author[1]{\textbf{Yichen Tang}}
\author[1]{\textbf{Yiqun Liu}}

\affil[1]{Department of Computer Science and Technology, Tsinghua University}
\affil[2]{Quan Cheng Laboratory}

\begin{document}
\maketitle

\begin{abstract}

Knowledge Editing is a technique that updates large language models (LLMs) with new information to maintain their world knowledge. 
This approach avoids the need to rebuild the model from scratch, thereby addressing the high costs associated with frequent retraining.
Among these, the in-context editing paradigm stands out for its effectiveness in integrating new knowledge while preserving the model's original capabilities. 
Despite its potential, existing in-context knowledge editing methods are often task-specific, focusing primarily on multi-hop QA tasks using structured knowledge triples. 
Moreover, their reliance on few-shot prompting for task decomposition makes them unstable and less effective in generalizing across diverse tasks.
In response to these limitations, we propose \textbf{EditCoT}, a novel knowledge editing framework that flexibly and efficiently updates LLMs across various tasks without retraining.
EditCoT works by generating a chain-of-thought (CoT) for a given input and then iteratively refining this CoT process using a CoT editor based on updated knowledge. 
We evaluate EditCoT across a diverse range of benchmarks, covering multiple languages and tasks. 
The results demonstrate that our approach achieves state-of-the-art performance while offering superior generalization, effectiveness, and stability compared to existing methods, marking a significant advancement in the field of knowledge updating\footnote{The code and data of EditCoT are available at: https://github.com/bebr2/EditCoT}.
\end{abstract}

\section{Introduction}

Large Language Models (LLMs) have shown remarkable performance across numerous NLP tasks in recent years.
However, once an LLM has been fully trained, its parametric knowledge becomes fixed and constrained.
Retraining the entire model to incorporate new information is both expensive and time-consuming. 
To address this challenge, the field of knowledge editing has emerged, offering techniques to modify or add knowledge efficiently while preserving most of the original model’s parameters.
Existing knowledge edit approaches can be broadly categorized into two types: parametric and non-parametric \cite{wang2024lekubelegalknowledgeupdate}. 
Parametric methods directly modify the model’s weights to integrate new knowledge. 
While this approach enables precise editing, existing research indicates that such changes may negatively affect the model’s original performance and interfere with previously learned knowledge that should remain unchanged \cite{gu2024modeleditingharmsgeneral}. 
In contrast, non-parametric methods, such as in-context editing, are increasingly popular because they utilize the in-context learning ability of LLMs without modifying the original model weights, thereby preserving the model’s foundational capabilities and existing knowledge \cite{zheng-etal-2023-edit}.

Most existing in-context knowledge editing frameworks are designed for multi-hop question answering (MQA) tasks using knowledge triples as the primary knowledge representation. Researchers construct multi-hop questions by linking multiple triples to evaluate these methods' performance \cite{zhong2024mquakeassessingknowledgeediting}. For instance, Mello and PokeMQA guide LLMs in task decomposition with manually crafted examples, facilitating knowledge updates for sub-questions \cite{zhong2024mquakeassessingknowledgeediting, gu2024pokemqaprogrammableknowledgeediting}. RAE \cite{shi2024retrievalenhancedknowledgeeditinglanguage} constructs and then edits a knowledge graph, utilizing the retrieval and pruning of the knowledge graph to obtain contextually injected knowledge.

While effective for triple-based MQA tasks, these methods are overly specialized and struggle with other tasks and complex reasoning. For example, RAE is impractical for the LeKUBE\cite{wang2024lekubelegalknowledgeupdate} legal dataset, where constructing a knowledge graph on its corpus is unfeasible. 
PokeMQA's strategy of using updated knowledge directly as answers to sub-questions works well for triple-based tasks but fails with more complex tasks like DUNE \cite{akyürek2023dunedatasetunifiedediting}. Furthermore, Mello and PokeMQA's reliance on few-shot prompting for task decomposition results in instability and reduce effectiveness across diverse tasks.

To address the limitations of existing in-context editing methods, we propose \textbf{EditCoT}, a framework that edits LLM's knowledge through chain-of-thought (CoT).
EditCoT is also an in-context editing method that does not change the parameters of LLMs used for reasoning, thereby preserving the original capacity of the backbone LLMs.
Unlike previous methods that are task-specific or heavily reliant on knowledge triples, EditCoT iteratively refines the model’s reasoning with retrieved updated knowledge. This flexible approach allows it to dynamically adapt its reasoning without requiring task-specific adjustments or predefined knowledge structures. Specifically, EditCoT starts by instructing the LLM to first generate an answer to the question, and then construct an initial CoT based on the question and answer.
The original CoT is then iteratively refined by a CoT editor, which requires only a single training session without needing retraining when encountering new knowledge.
The final answer is generated by prompting the LLM to inference based on the updated CoT.

We evaluate EditCoT on a variety of benchmarks, including the triple-based multi-hop QA dataset MQuAKE-CF-3k\cite{zhong2024mquakeassessingknowledgeediting}, a free-form editing task dataset DUNE\cite{akyürek2023dunedatasetunifiedediting}, and a domain-specific Chinese legal knowledge update dataset LeKUBE\cite{wang2024lekubelegalknowledgeupdate}. 
They cover different editing formats and tasks in both Chinese and English. 
Experimental results demonstrate that our method achieves state-of-the-art performance with better efficiency and robustness.

The contributions of this paper are as follows: 
\vspace{-1mm}
\begin{itemize}[leftmargin=0.015\textwidth]
    \item We propose a novel framework, \textit{EditCoT}, that edits the chain-of-thought of LLMs, enabling more effective and task-agnostic knowledge updates.
    
    \vspace{-0.5mm}
    \item We demonstrate our method performs well across various knowledge editing benchmarks with different editing formats and task types, exhibiting significantly better effectiveness, robustness, and generalizability compared to other baselines.

\end{itemize}

\section{Related Work}

\subsection{Knowledge Editing}

Knowledge editing methods modify or update knowledge within LLMs, categorized into parametric and non-parametric approaches \cite{wang2024lekubelegalknowledgeupdate}. ROME \cite{meng2023locatingeditingfactualassociations} is a parametric method, update the model's parameters by treating FFN layers as key-value storage for precise modifications. Non-parametric approaches leverage in-context learning \cite{brown2020languagemodelsfewshotlearners}, such as Mello \cite{zhong2024mquakeassessingknowledgeediting}, which uses prompts for task decomposition, PokeMQA \cite{gu2024pokemqaprogrammableknowledgeediting}, which enhances Mello's robustness with a scope detector, and RAE \cite{shi2024retrievalenhancedknowledgeeditinglanguage}, which uses a knowledge graph for task-specific editing.

\subsection{Chain-of-Thought}

Chain-of-thought (CoT) has significantly enhanced LLMs by providing step-by-step reasoning for complex tasks. ~\citet{wei2023chainofthoughtpromptingelicitsreasoning} introduce CoT prompting, where manually supplied reasoning chains help LLMs generate multi-step solutions. ~\citet{kojima2023largelanguagemodelszeroshot} demonstrate phrases like "Let's think step by step" enable zero-shot CoT reasoning.
Recent work has focused on enhancing the factuality of CoT. ~\citet{wang2023selfconsistencyimproveschainthought} propose a self-consistency decoding method to improve CoT reliability. \citet{zhao2023verifyandeditknowledgeenhancedchainofthoughtframework} introduce Verify-and-Edit that involves editing the generated CoT. However, our approach significantly differs in some key aspects. The verification process of Verify-and-Edit relies on self-consistency, which is less effective for knowledge editing due to the LLM's high confidence in outdated information. Additionally, their method depends on the LLM's ability to learn from context, whereas our CoT editor provides more comprehensive adjustments. Further comparative analysis can be found in ~\autoref{cha:vae}.

\section{Methodology}
In this section, we present the methodology of the EditCoT framework in detail. 
We begin with an overview of our proposed approach, followed by a step-by-step description of the framework’s workflow.
We then explain the CoT editor training process in depth, concluding with a discussion of implementation details and dataset construction.

\subsection{Methodology Overview}

\begin{figure}
    \centering
    \includegraphics[width=\linewidth]{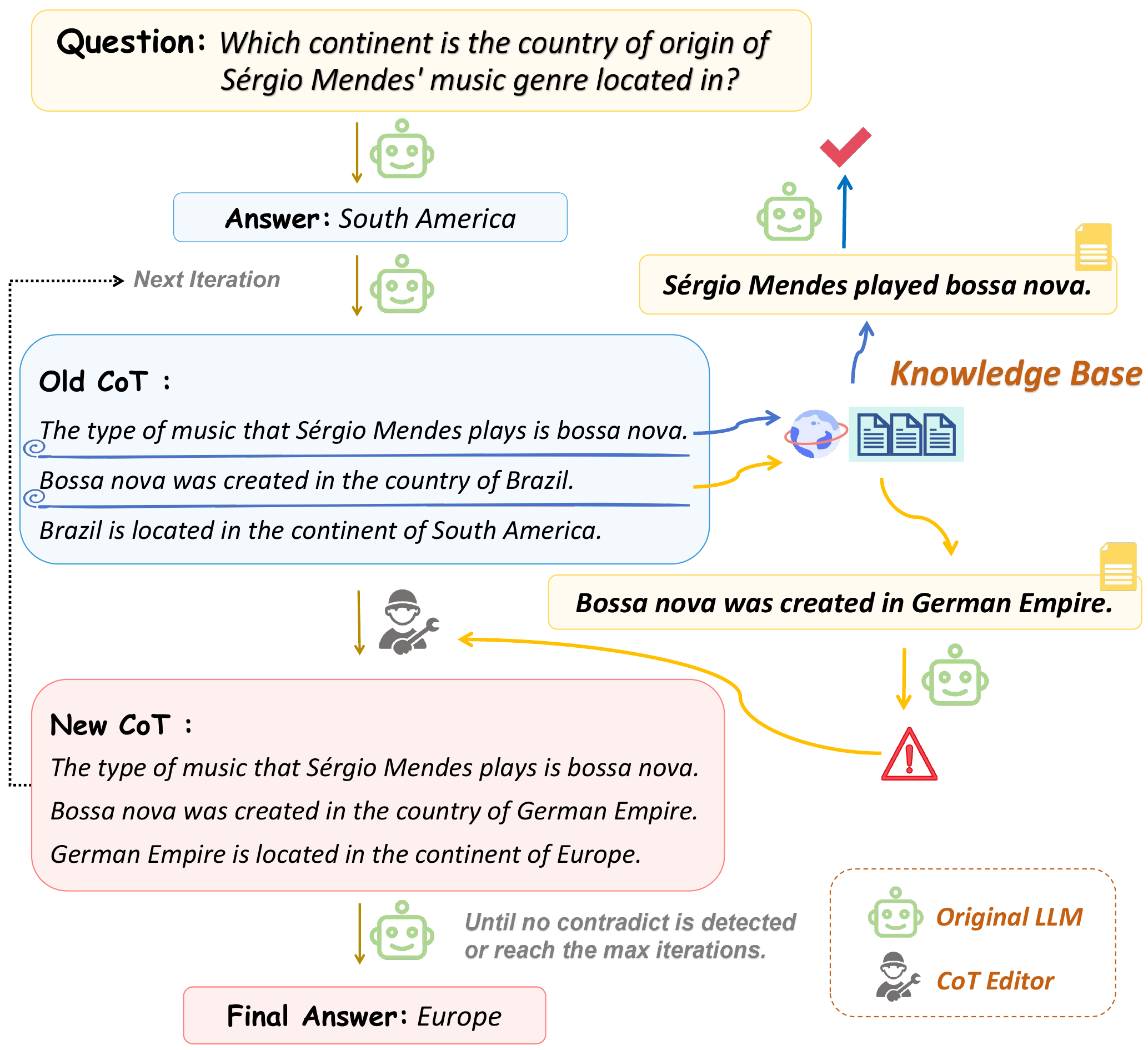}
    \vspace{-2mm}
    \caption{EditCoT updates LLMs by iteratively editing the CoT to integrate new knowledge while preserving its reasoning abilities. It generates an initial CoT, detects conflicts, and revises the CoT iteratively.}
    \vspace{-2mm}
    \label{fig:editcotframework}
\end{figure}


EditCoT aims to update LLMs by editing their CoT, integrating new factual knowledge while preserving their reasoning capabilities.
Traditional parametric knowledge editing methods often require resource-intensive retraining or fine-tuning, which can result in the loss of existing knowledge.
In contrast, EditCoT introduces an iterative editing mechanism that efficiently incorporates updates by focusing solely on the CoT, ensuring that the model remains both accurate and logically consistent.

The framework leverages two components: an unedited version of the LLM responsible for generating the initial CoT, detecting conflict, and generating the final answer, and an editor capable of editing CoT based on the new knowledge. 
The following subsection details each step of EditCoT.

\subsection{Workflow of EditCoT}

\autoref{fig:editcotframework} illustrates the EditCoT framework’s overall workflow, which is also described in pseudocode in \autoref{apd:pse}. The core steps are as follows:

\paragraph{Step 1: Generate Initial CoT by the Original LLM.} 
At the beginning of the process, the unedited LLM generates an initial CoT in response to a query. This CoT outlines the sequence of reasoning steps leading to the model's final answer. Prioritizing the completeness of the reasoning path over the answer's accuracy, we first let the model provide the answer directly and then generate the CoT based on the question and answer.


\paragraph{Step 2: Retrieve Step by Step and Detect Conflicts.} 
\label{cha:retrieve}
Once the initial CoT is generated, we treat each step of the CoT as a query and retrieve it from the memory of edited knowledge. This memory is typically organized in two ways: one where the pre- and post-edit knowledge are paired together (e.g., MQuAKE \cite{zhong2024mquakeassessingknowledgeediting}), and another where only the post-edit knowledge is provided (e.g., DUNE \cite{akyürek2023dunedatasetunifiedediting}).
For the former, we search the knowledge base for the pre-edit knowledge and then match the post-edit knowledge according to the corresponding relationship. For the latter, we directly search within the post-edit knowledge. Once updated knowledge is found, we proceed to the conflict detection stage.

EditCoT utilizes the original LLM to detect knowledge conflicts by engaging in a three-class classification task: conflict, support, or irrelevant. Specifically, given a question, the initial CoT, and the retrieved knowledge, the classification is determined based on the probability of the prefix of the corresponding word output by the LLM. This process is crucial for identifying whether the original reasoning diverges from updated knowledge. 

\paragraph{Step 3: Edit the chain-of-thought}
Once a knowledge conflict is detected, we input the question, new knowledge, and old CoT into a CoT editor, which generates a new CoT. Then, our iterative process returns to Step 2, where this new CoT is treated as the old CoT for the next iteration. Each iteration resolves one edit at a time, progressively addressing multiple potential conflicts through iterations. The process stops when all conflicts are found or the iteration limit is reached.

\paragraph{Step 4: Get Final Answer}
Once the CoT has been fully edited and no further conflicts are detected, the final chain-of-thought is used to generate the final answer. The original LLM is responsible for this step, ensuring that the model's answer adheres to its existing inferential abilities, while also incorporating the new factual knowledge. We instruct the original LLM to pay more attention to the new CoT in the context since the new CoT often conflicts with the model's intrinsic knowledge.

\subsection{CoT Editor Training}

\begin{figure*}
    \centering
    \includegraphics[width=0.9\linewidth]{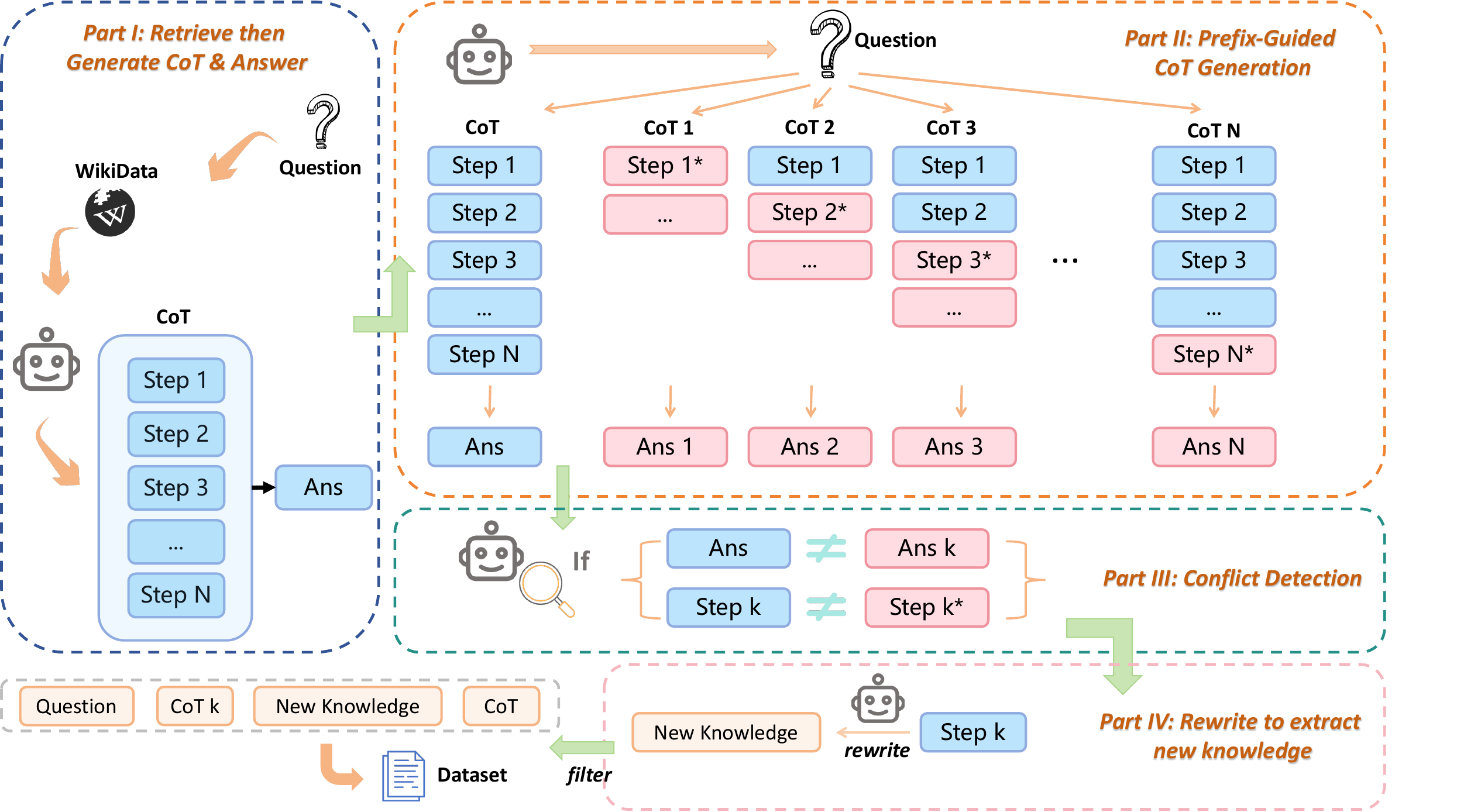}
    \vspace{-2mm}
    \caption{Diagram of the dataset construction: Our data source consists solely of questions from HotpotQA, without requiring human-annotated answers or related articles. In Part II, the blue sections represent the prefix used to guide generation, which have been generated in Part I, while red sections indicate newly generated reasoning paths.}
    \label{fig:dataconstruct}
    \vspace{-3mm}
\end{figure*}

We design an automated approach to generate training data required for the CoT editor, where human annotation is unnecessary, except for the use of some human-posed questions. 
Our method does not rely on state-of-the-art LLMs like GPT-4\cite{openai2024gpt4technicalreport} and is not tailored to specific datasets or tasks. The editor learns to edit CoT during training and can apply this skill to various tasks without retraining when encountering new knowledge. Additionally, the CoT editor is trained from the original LLM, ensuring that no knowledge distillation issues arise.


\subsubsection{Task Formulation}
The editor's input includes a question $Q$, an original $\text{CoT}$, and new knowledge $K_{\text{new}}$ that conflicts with the CoT. The goal is to generate a revised chain-of-thought $\text{CoT}_{\text{new}}$ that incorporates new knowledge. Formally, this can be expressed as:

\vspace{-1em}
\begin{small}
\begin{align}
\text{Editor}(Q, \text{CoT}, K_{\text{new}}) \rightarrow \text{CoT}_{\text{new}} \nonumber,
\end{align}
\end{small}

\subsubsection{Data Construction}
To construct training data, we start by collecting high-quality, human-generated questions. In practice, we use the validation and test sets from the HotpotQA\cite{yang2018hotpotqadatasetdiverseexplainable} dataset, which contains diverse multi-hop questions that often require multiple Wikipedia articles to answer. HotpotQA is selected for its diversity of question types, and multi-hop reasoning tasks generally yield more complex CoTs. After obtaining these questions, we use the following construction steps:

\paragraph{Part I: Retrieve, then Generate CoT and Answer}

For each question, we first retrieve top-5 relevant Wikipedia paragraphs and then use a retrieval-augmented generation (RAG) method to generate the CoT and the corresponding answer. We consider this initial CoT and answer as a basic version, and in the following sections, these are highlighted as {\colorbox {light-blue}{CoT}} and  {\colorbox {light-blue}{Answer}} for clarity. To facilitate step-level separation, we prompt the model to indicate each step with the \texttt{[STEP]} phrase.

\paragraph{Part II: Prefix-Guided CoT Generation}
Given a CoT with $N$ steps generated in Part I, for each step, e.g. the kth step ($k=1, 2, \dots, N$), we take the first $(k-1)$ steps as a prefix and prompt the model to generate the remaining steps and the final answer, marked as {\colorbox {light-pink}{CoT k}} and {\colorbox {light-pink}{Answer k}} respectively. It is important to note that this step does not use RAG, and the model is only prompted with the first $(k-1)$ steps of the CoT as context. We finally obtain $N$ CoTs in this part.

\paragraph{Part III: Conflict Detection}
We then compare each of the $N$ CoTs generated in Part II to the original {\colorbox {light-blue}{CoT}} and instruct the model to detect any knowledge conflicts. A data point is accepted into the final dataset only if: (1) there is a conflict in the final answers, and (2) there is a conflict at the first diverging step between the two CoTs.

For example, considering  {\colorbox {light-pink}{CoT k}} and {\colorbox {light-blue}{CoT}}, the first diverging step is step k. If the original  {\colorbox {light-blue}{Step k}} and the new {\colorbox {light-pink}{Step k}} conflict with each other, and there is also a conflict between {\colorbox {light-blue}{Answer}} and {\colorbox {light-pink}{Answer k}}, they are eligible to be added to the final dataset. This ensures that conflicting knowledge can be pinpointed at a specific step in the reasoning chain. Here we assume that the divergence occurs only once in most of the cases.

\paragraph{Part IV: Rewrite to Extract New Knowledge}
In this step, we rewrite the identified diverging step from the original {\colorbox {light-blue}{CoT}} to explicitly extract the new knowledge. Specifically, for the example above, a complete training data entry is formed as a tuple:

\vspace{-1em}
\begin{small}
\begin{align}
& (Q, \enspace \text{CoT}, \enspace K_{\text{new}}, \enspace \text{CoT}_{\text{new}}) = \nonumber \\
& (\text{Question}, \enspace \text{\colorbox {light-pink}{CoT k}}, \enspace \text{rewrite({\colorbox {light-blue}{Step k}})}, \enspace \text{\colorbox {light-blue}{CoT}}) \nonumber.
\end{align}
\end{small}
We also apply regular expressions to filter out low-quality examples, such as those overly influenced by the retrieved documents in RAG. ~\autoref{fig:dataconstruct} illustrates the overall process of dataset construction.

\subsection{Implementation Details for Training}
\label{cha:dataconstruct}
\begin{figure}[t]
    \centering
    \includegraphics[width=\linewidth]{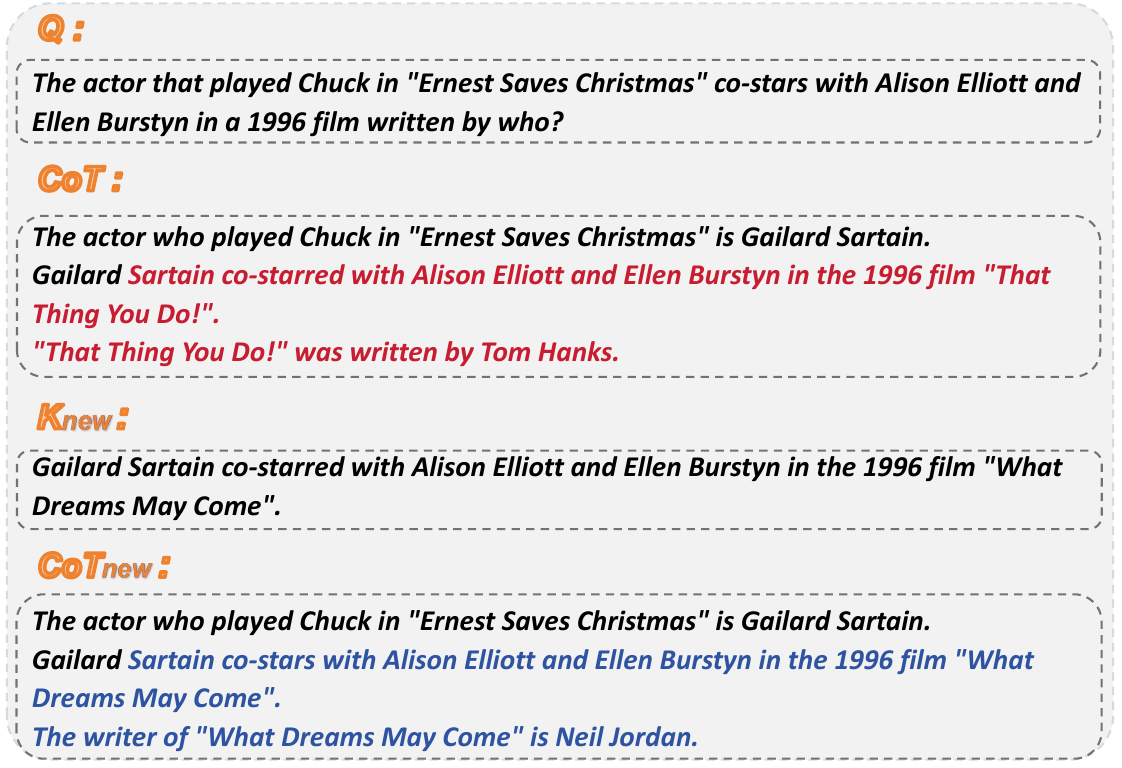}
    \caption{A training data example. The black part of the two CoTs indicates a shared prefix.}
    \label{fig:de}
\end{figure}

We implement conflict detection using a multiple-choice format. If the model assigns a probability of 0.6 or higher to a conflict between two answers or reasoning steps, we consider them inconsistent.
During prefix-guided generation, to encourage the LLM to focus on the same topic at the diverging step, we provide the first word of the original CoT at the diverging step to the LLM. Prompt templates for dataset construction are detailed in \autoref{apd:prompt1}.
The dev and test sets of HotpotQA contain 14,810 questions. Using Meta-Llama-3-8B-Instruct\cite{dubey2024llama3herdmodels} and Qwen2.5-14B-Instruct\cite{qwen2, qwen2.5}, we generate 459 and 316 training examples, respectively. \autoref{fig:de} illustrates a constructed data instance. During training, we calculate the loss based on $CoT_{new}$ using cross-entropy loss.

\section{Experimental Settings}

\subsection{Dataset and Metrics}

\label{cha:dataset}

We evaluate the performance of EditCoT and other baselines across the following three datasets:

\paragraph{MQuAKE-CF-3k\cite{zhong2024mquakeassessingknowledgeediting}}
This dataset is designed for knowledge editing using Wikipedia-based knowledge triples, with queries in the form of multi-hop questions. Each question involves 1-4 edits. The evaluation metric is \textit{Multihop-Accuracy}, as in the original paper: each instance contains 3 questions, and if anyone is answered correctly, it is considered accurate. We follow the settings of Mello and PokeMQA, using \textit{Exact Match} to assess correctness.

\paragraph{DUNE\cite{akyürek2023dunedatasetunifiedediting}}
It includes various editing tasks, where edits are natural language instructions for the model to follow specific facts or requirements. It covers correcting errors, adding new information, etc. We conduct evaluations on 3 subsets: \textit{Arithmetic Reasoning}, \textit{Scientific Reasoning}, and \textit{New Information}. The metric is \textit{Accuracy}.

\paragraph{LeKUBE\cite{wang2024lekube}}
A Chinese legal knowledge editing dataset . We focus on two challenging tasks, where prior baselines show poor performance: \textit{Multiple-Choice Questions of the Legal Scenario} and \textit{True/False Questions of Change in Statute}. The metric is \textit{Accuracy}. The statistics of the three benchmarks are detailed in \autoref{apd:benchmark}.

\subsection{Models}
We evaluate two open-source models: Meta-Llama-3-8B-Instruct\cite{dubey2024llama3herdmodels} and Qwen2.5-14B-Instruct\cite{qwen2.5}. Since the former does not support Chinese, we evaluate the latter on LeKUBE. Following LeKUBE's protocol, we first fine-tune the model on the STARD corpus\cite{su2024stardchinesestatuteretrieval}, which contains Chinese laws and judicial interpretations, to ensure the model acquires Chinese legal knowledge before any updates.

We also assess GPT-4o-2024-08-06\cite{openai2024gpt4technicalreport} on MQuAKE-3k-CF. Due to its proprietary nature, we test only a few non-parametric editing methods on this model. Following RAE\cite{shi2024retrievalenhancedknowledgeeditinglanguage}, we randomly sample 300 data points to minimize costs. Meta-Llama-3-8B-Instruct is used as a proxy model for tasks the proprietary model cannot perform. In RAE, this proxy extracts and prunes knowledge graphs based on probabilities, while in EditCoT, it acts as a CoT editor.

\subsection{Baselines}
We compare EditCoT with a range of model editing techniques, including knowledge neuron-based (KN)\cite{dai2021knowledge}, rank-one model editing (ROME)\cite{meng2023locatingeditingfactualassociations}, and in-context methods such as Mello\cite{zhong2024mquakeassessingknowledgeediting}, PokeMQA\cite{gu2024pokemqaprogrammableknowledgeediting}, and RAE\cite{shi2024retrievalenhancedknowledgeeditinglanguage}. We also include full-parameter fine-tuning (FT) and retrieval-augmented generation (Naive RAG)\cite{lewis2021retrievalaugmentedgenerationknowledgeintensivenlp} as baselines. To distinguish our approach from the RAG method, we evaluate two advanced variants, RAT\cite{wang2024ratretrievalaugmentedthoughts} and FLARE\cite{jiang2023flare}, on the MQuAKE and DUNE benchmarks. RAT incorporates multi-turn query rewriting, retrieval, and CoT enhancements to address complex tasks effectively. FLARE dynamically determines the necessity of retrieval during model inference. These comparisons highlight the superiority of our approach in performing knowledge editing tasks.

Notably, since LeKUBE is a Chinese dataset, we translate the official prompts for Mello and PokeMQA into Chinese to accommodate their sensitivity to language type. RAT and FLARE are not evaluated on LeKUBE because it is challenging to align their settings with those of the dataset. RAE is not applicable to datasets other than MQuAKE-CF-3k since it requires construction of knowledge graphs, which is impractical on these datasets. Details on the baselines are available in \autoref{apd:baselines}.

\begin{table*}[t]
\caption{Results on MQuAKE-CF-3k and three subsets of DUNE, with MQuAKE-CF-3k using the Multi-hop Accuracy metric and the others using Accuracy. The best performance for each dataset and model is highlighted in bold, and the second-best is underlined. A dash ("-") indicates that the method is not applicable to the given dataset. Llama and Qwen represent Llama-3-8B-Instruct and Qwen2.5-14B-Instruct, respectively.}
\vspace{-1mm}
\renewcommand{\arraystretch}{1.2}
\resizebox{\linewidth}{!}{
\begin{tabular}{ccccccccccccc}
\hline
\multirow{2}{*}{\textbf{Models}}                                   & \multicolumn{2}{c}{\multirow{2}{*}{\textbf{Datasets}}} & \multicolumn{3}{c|}{\textbf{Parametric Methods}}                        & \multicolumn{3}{c|}{\textbf{RAG Methods}}                               & \multicolumn{4}{c}{\textbf{In-Context Editing}}                            \\ \cline{4-13} 
                                                                   & \multicolumn{2}{c}{}                                   & \textbf{FT} & \textbf{KN} & \multicolumn{1}{c|}{\textbf{ROME}} & \textbf{Naive RAG} & \textbf{RAT} & \multicolumn{1}{c|}{\textbf{FLARE}} & \textbf{Mello} & \textbf{PokeMQA} & \textbf{RAE}  & \textbf{EditCoT(Ours)} \\ \hline
\multirow{4}{*}{\textbf{Llama}}                      & \multicolumn{2}{c}{\textbf{MQuAKE-CF-3k}}              & 11.2                 & 2.8         & 4.0                                & 9.7                & 0.7          & 3.3                                & 10.0           & 26.1             & \textbf{54.1} & {\ul 35.4}             \\ \cline{2-3}
                                                                   & \multirow{3}{*}{\textbf{DUNE}}  & \textbf{Arithmetic}  & 74.1                 & 80.4        & 80.5                               & {\ul 84.0}         & 70.0         & 40.1                                & 73.7           & 83.8             & -             & \textbf{90.7}          \\
                                                                   &                                 & \textbf{Scientific}  & 11.1                 & 82.1        & 81.4                               & 81.6               & {\ul 84.3}   & 82.1                                & 77.1           & 61.7             & -             & \textbf{85.0}          \\
                                                                   &                                 & \textbf{New Info}    & 30.3                 & 66.6        & 69.1                               & 89.2               & 77.0         & 81.6                                & {\ul 89.8}     & 58.7             & -             & \textbf{91.3}          \\ \hline
\multicolumn{1}{l}{\multirow{4}{*}{\textbf{Qwen}}} & \multicolumn{2}{c}{\textbf{MQuAKE-CF-3k}}              & 9.3                  & 3.2         & 0.0                                & 10.0               & 3.3          & 3.7                                 & 5.8            & 5.3              & {\ul 26.9}    & \textbf{34.2}          \\ \cline{2-3}
\multicolumn{1}{l}{}                                               & \multirow{3}{*}{\textbf{DUNE}}  & \textbf{Arithmetic}  & 82.8                 & 86.9        & 86.9                               & {\ul 89.7}         & 82.6         & 23.4                                & 61.0           & 35.2             & -             & \textbf{97.3}          \\
\multicolumn{1}{l}{}                                               &                                 & \textbf{Scientific}  & 13.8                 & 75.3        & 75.5                               & 70.7               & {\ul 86.0}   & 83.2                                & 37.0           & 7.7              & -             & \textbf{86.3}          \\
\multicolumn{1}{l}{}                                               &                                 & \textbf{New Info}    & 76.2                 & 73.4        & 73.1                               & \textbf{96.8}      & 87.4         & 70.1                                & 34.3           & 17.3             & -             & {\ul 93.8}             \\ \hline
\end{tabular}
}
\vspace{-1mm}
\label{tab:mainresult}
\end{table*}

\subsection{Implementation Details}

For the CoT editor, the data construction process is outlined in \autoref{cha:dataconstruct}. We use 80\% of the filtered data generated by the respective model and train it for 1 epoch. To generate the initial answer, CoT, and final answer, and detect conflicts, we use 5-shot prompts styled after Mello and PokeMQA and apply this format across all tasks, as detailed in \autoref{apd:prompt2} . For conflict detection, we instruct the model to classify outputs into three categories: "Contradict", "Support", or "Unrelated". We take the token with the highest probability as the final classification. While we use the \texttt{[STEP]} token during dataset construction, in actual inference, we allow the model to freely generate the CoT, with steps separated by newline characters.

For the EditCoT retrieval corpus, settings are provided in \autoref{cha:retrieve}. For datasets offering both pre- and post-edit knowledge (e.g., MQuAKE-CF-3k and LeKUBE), we retrieve from the pre-edit corpus and match with corresponding post-edit sentences. For datasets with only post-edit knowledge (e.g., DUNE), we directly retrieve from the post-edit corpus.
We use the same retriever, Contriever\cite{izacard2022unsuperviseddenseinformationretrieval}, as Mello in MQuAKE-CF and DUNE. For Chinese datasets, we use BM25\cite{robertson2009probabilistic}, which performs well in LeKUBE.

\section{Experimental Results}

\begin{table*}[t]
\centering
\caption{Applicability to vertical domains: LeKUBE is a Chinese Legal Knowledge Editing Benchmark.}
\vspace{-1mm}
\renewcommand{\arraystretch}{1.1}
\resizebox{\linewidth}{!}
{
\begin{tabular}{llccccccc}
\hline
\textbf{Model} &
  \textbf{Subsets of LeKUBE} &
  \textbf{Fine-tuning} &
  \textbf{Naive RAG} &
  \textbf{KN} &
  \textbf{ROME} &
  \textbf{Mello} &
  \textbf{PokeMQA} &
  \textbf{EditCoT(Ours)} \\ \hline
\multirow{2}{*}{\textbf{Qwen2.5-14B-Instruct}} &
  \textbf{MCQ of the Legal Scenario} &
  42.2 &
  {\ul 47.2} &
  38.9 &
  38.3 &
  6.7 &
  0.0 &
  \textbf{58.3} \\
 &
  \textbf{T/F Questions of Change in Statute} &
  65.9 &
  {\ul 69.2} &
  55.8 &
  56.1 &
  18.6 &
  0.0 &
  \textbf{69.5} \\ \hline
\end{tabular}

}
\label{tab:lekube}
\end{table*}


\subsection{Effectiveness}
\label{cha:mainrag}

\begin{table}[]
\centering
\caption{Results on GPT-4o (a black-box LLM). Llama-3-8B-Instruct serves as a proxy LLM for RAE and EditCoT. The dataset is 300 samples from MQuAKE-CF.}

\resizebox{\linewidth}{!}{
\begin{tabular}{cccccc}
\toprule
               & \textbf{Naive RAG} & \textbf{Mello} & \textbf{PokeMQA} & \textbf{RAE}  & \textbf{EditCoT(Ours)} \\ \toprule
\textbf{Score} & 12.0         & 15.0           & 9.7              & \textbf{59.7} & {\ul 45.0}             \\ \bottomrule
\end{tabular}
}
\label{tab:gpt4oresult}
\end{table}

\textbf{Generalizability.} EditCoT demonstrates superior dataset generalization and model robustness compared to baselines. As shown in \autoref{tab:mainresult}, EditCoT achieves top performance on 6 out of 8 results and second-best on 2 out of 8 results in MQuAKE-3k-CF and DUNE, outperforming all baselines. It consistently excels across two LLMs, unlike other methods like Mello, PokeMQA, and RAE, which perform well on Llama-3-8B-Instruct but poorly on Qwen2.5-14B-Instruct. Mello and PokeMQA's lack of robustness stems from their dependence on strict instructions and hand-crafted few-shot examples, which exhibit varying effectiveness across models and tasks. Additionally, safety-aligned models often reject external edits. To provide a detailed comparison between our method and the baselines, we include case studies in \autoref{apd:cs}.

\textbf{Applicability to Vertical Domains and Black-box Models.} EditCoT is also effective in specialized vertical domains and proprietary models. \autoref{tab:lekube} presents the evaluation results on LeKUBE, where EditCoT outperforms all baselines across two tasks, significantly surpassing other methods in the multiple-choice questions of the legal scenario. Furthermore, \autoref{tab:gpt4oresult} reports GPT-4o's performance on MQuAKE-CF, where EditCoT also demonstrates competitive results, achieving second-best overall. Notably, the proxy model in this experiment is Meta-Llama-3-8B-Instruct. When we compare the performance of the proxy model itself with the GPT-4o, we find that, while RAE's accuracy improves from 54.1\% to 59.7\% with GPT-4o, EditCoT achieves a larger relative improvement, rising from 35.4\% to 45.0\%.
This suggests that EditCoT relies more on the tested LLM’s internal reasoning, while the CoT Editor (8B size) emphasizes editing over reasoning. And RAE benefits primarily from leveraging external knowledge.

\textbf{Challenges in General-Purpose In-Context Editing.} Developing a broadly applicable in-context editing framework remains a significant challenge across both datasets and models. The three datasets span multihop reasoning, generalized edits, domain-specific knowledge, and different languages, posing difficulty for current editing methods. Among the baselines, PokeMQA performs well on multi-hop questions with Llama-3, and simpler methods like Naive RAG excel in narrow settings. However, these methods perform poorly on other datasets or models. RAE excels on the MQuAKE-CF-3k, because it constructs structured knowledge graphs and accesses a broader range of external knowledge (e.g., Wikipedia), but fails to generalize beyond structured datasets. In contrast, EditCoT addresses these limitations, achieving the best or second-best performance across datasets and models, although not always optimal.

\textbf{RAG methods underperform compared to EditCoT}. There are two reasons: 1) RAG methods seldom address conflicts between parametric and contextual knowledge, which are common in knowledge editing. While RAT modifies CoT, \autoref{tab:abl} shows that EditCoT, using a trained editor, injects knowledge more effectively than a simple prompt. 2) Certain assumptions of RAG are incompatible with knowledge editing. FLARE dynamically decides on retrieval based on model uncertainty about generated words. However, in knowledge editing, the LLM is often confident about the outdated knowledge. In \autoref{cha:rag}, we provide a more detailed discussion on the differences and advantages of EditCoT relative to the RAG methods.

\subsection{Robustness}

    
In this section, we evaluate EditCoT against other in-context knowledge editing methods, using Llama3-8B-Instruct from multiple perspectives.

\textbf{Locality:}
Locality in knowledge editing assesses a method's ability to update specific knowledge without impairing untargeted information \cite{mitchell2022memorybasedmodeleditingscale}. We evaluate locality using the New Information (Locality) subset of DUNE, where the retrieval scope is limited to the edited set (consistent with the main experimental setup), but test tasks pertain to knowledge outside this set. The results, presented in \autoref{tab:locality}, are compared with the unedited model's performance. While all editing methods show some decline in performance relative to the unedited model, EditCoT exhibits the best locality. RAG performs the worst, as it lacks conflict detection between documents and queries, resulting in poor handling of irrelevant information.
\begin{table}[t]
\caption{Locality test on the New Info (Locality) set of DUNE. EditCoT is the best in all methods.}
\vspace{-1mm}
\resizebox{\linewidth}{!}{
\begin{tabular}{c|c|cccc}
\toprule
                  & \textbf{Before Editing} & \textbf{Naive RAG} & \textbf{Mello} & \textbf{PokeMQA} & \textbf{EditCoT} \\ \midrule
\textbf{Accuracy} & 65.2                    & 34.3         & 58.0     & 45.6             & \textbf{59.7}    \\ 
\bottomrule
\end{tabular}
}
\vspace{-1mm}
\label{tab:locality}
\end{table}

\begin{figure}[t]
    \centering
        \includegraphics[width=0.9\linewidth]{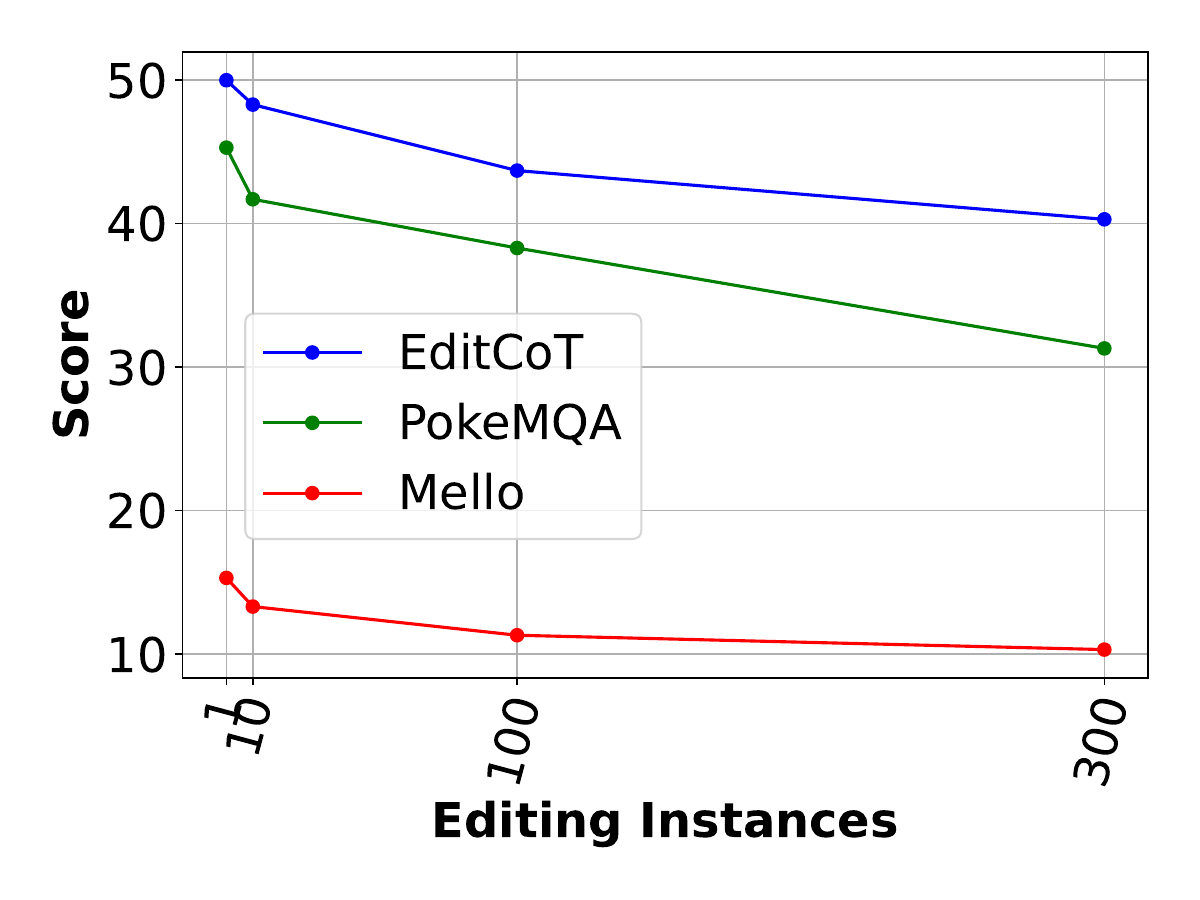}
        \vspace{-5mm}
        \caption{Performance with different batch sizes on MQuAKE-CF. The metrics here is multi-hop accuracy.}
        \label{fig:batchsize}
\end{figure}

\textbf{Performance with Different Batch Sizes:}
The performance of knowledge editing methods varies with the editing batch size. Here we denote the batch size as the number of questions in one batch. We test the sensitivity of Mello, PokeMQA, and EditCoT to batch sizes.\footnote{We don't compare RAE because the locality of in-context editing is influenced by retrieval. RAE uses knowledge graph retrieval, which is not comparable to the other three methods.} We evaluate batch sizes of 1, 10, 100, and 300 on a 300-sample dataset from MQuAKE-CF-3k. Results in \autoref{fig:batchsize} indicate that all methods perform best with a single sample, with performance decreasing as size increases. However, EditCoT consistently outperforms the others across all batch sizes, showing the smallest performance degradation (around 20\%) from size 1 to 300, compared to declines of over 30\% for the others.

\subsection{Efficiency}

\begin{figure}[t]
        \centering
        \includegraphics[width=0.9\linewidth]{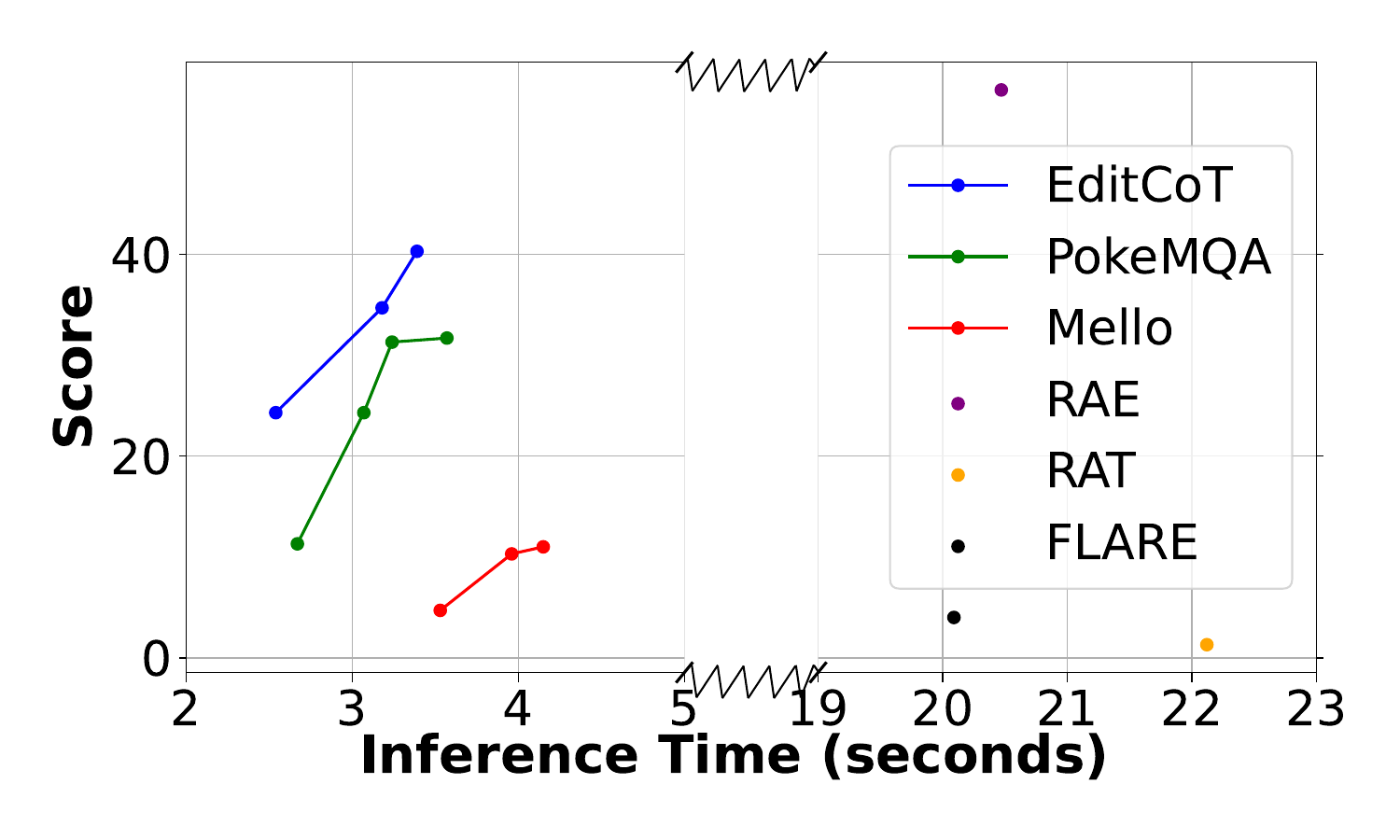}
        \vspace{-3mm}
        \caption{Inference time vs. performance: Top-left indicates shorter inference time and higher performance. Figure split due to wide horizontal span.}
        \label{fig:time}
\end{figure}

We investigate efficiency on 100 sampled questions from MQuAKE-CF-3k. By varying the max-iterations for Mello, PokeMQA, and EditCoT, we record their corresponding inference time and performance. Complete experimental details are in \autoref{apd:efficiency}. As shown in \autoref{fig:time}, EditCoT achieves superior performance with the shortest inference time, outperforming baselines at similar computational costs. The efficiency advantage stems from requiring fewer iterations (3.65 average iterations for EditCoT vs. 4.33 average iterations for PokeMQA) and demonstrating stronger early performance within initial iterations. Further details on the impact of iteration limits are in \autoref{apd:maxiter}. The average inference time for RAE and two advanced RAG methods exceeds 20 seconds, making their application in real-world tasks challenging.

\subsection{Ablation Study}

\begin{table}[t]
    \centering
    \caption{Impact of conflict detection and trained CoT editor. L-8B represents  Llama-3-8B-Instruct and Q-14B represents Qwen2.5-14B-Instruct.}
    \resizebox{\linewidth}{!}{
    \begin{tabular}{lllc}
\toprule
                                       & \multicolumn{2}{c}{\textbf{MQuAKE-CF}}                                 & \textbf{MCQ in LeKUBE} \\ \cline{2-4} 
                                       & \multicolumn{1}{c}{\textbf{L-8B}} & \multicolumn{1}{c}{\textbf{Q-14B}} & \textbf{Q-14B}         \\ \midrule
\textbf{EditCoT}                       & 40.3                              & 40.3                               & 58.3                   \\
\textbf{EditCoT w/o conflict detection} & 32.3 & 33.0 &  48.9 \\
\textbf{EditCoT w/o trained CoT editor} & 29.7                              & 26.7                               & 52.2                   \\ \bottomrule
\end{tabular}
}
\label{tab:abl}
\end{table}

\begin{table}[t]
\centering
\caption{Impact of training settings on the performance of the CoT editor, evaluated on MQuAKE-CF.}
\resizebox{\linewidth}{!}{
\begin{tabular}{cccccc}
\toprule
\textbf{\# Epoch}              & \textbf{1}  & \textbf{2}   & \textbf{3}   & \textbf{4}   & \textbf{5}   \\
\textbf{Accuracy} & 40.3 & \textbf{40.7} & 40.3          & 40.0 & 40.0 \\ \toprule
\textbf{\# Training Data} & \textbf{50} & \textbf{100} & \textbf{200} & \textbf{250} & \textbf{316(All)} \\
\textbf{Accuracy} & 39.0 & 39.3         & \textbf{40.7} & 40.3 & 40.3 \\ \bottomrule
\end{tabular}
}
\label{tab:abl2}
\end{table}

In this section, MQuAKE-CF means a dataset of 300 examples sampled from MQuAKE-CF-3k.

\label{cha:ic}
\paragraph{Impact of Components}
We evaluate two components of our method, as shown in \autoref{tab:abl}.
Removing conflict detection before editing the CoT or employing the original model as the CoT editor without specialized training result in considerable performance deterioration. Additionally, the degree of performance decline varies among different datasets when one component is removed.
These results demonstrate the critical role of both components: conflict detection prevents contamination from irrelevant retrieved information, while training the editor enables effective utilization of model editing capabilities beyond simple prompting.
Furthermore, in \autoref{app:sft_baseline_hotpotqa_investigation}, we examine whether training the CoT editor results in data leakage, ensuring the fairness of the experiment.


\paragraph{Impact of Training Configurations}
~\autoref{tab:abl2} illustrates the effect of training dataset sizes and epochs on the CoT editor. The model tested is Qwen2.5-14B-Instruct. While increasing the number of epochs has little impact, performance stabilizes after expanding the training size to 200 examples, suggesting that the CoT editor requires a sufficient amount of diverse data to learn effectively. This supports our main experimental setup, where training for one epoch with 80\% of generated data (252 examples) is a reasonable choice.

\section{Conclusion}

In this paper, we introduce \textbf{EditCoT}, a novel framework designed for efficiently updating LLMs through iterative CoT editing. EditCoT enhances the generalizability and robustness of knowledge editing across diverse tasks, without the need for retraining the LLM each time new knowledge is encountered. Our experiments, conducted on a variety of benchmarks, demonstrate that EditCoT consistently outperforms existing methods, proving its effectiveness in both general-purpose and domain-specific applications. This approach presents a flexible and efficient solution for continuous knowledge updates in LLMs, marking a significant advancement in the field of knowledge editing.


\section{Limitations}

EditCoT has certain limitations, particularly in its reliance on two distinct models. While this design choice does not increase inference time, it does result in higher GPU resource consumption, which can be challenging in environments with limited hardware resources. However, in practical applications, there is always a trade-off between performance and resource consumption. Compared to other in-context editing approaches, EditCoT offers notable improvements in both efficiency and performance (as shown in \autoref{fig:time}). As for resource consumption, the experiments in \autoref{tab:gpt4oresult} demonstrate that even with large-scale models like GPT-4o, employing an 8B CoT editor as a proxy model delivers strong performance while incurring minimal resource overhead relative to the base LLM being edited (GPT-4o). This implies that in real-world applications, the CoT Editor can be much smaller than the original LLM.

For the CoT editor, the extent to which the model size can be reduced and the potential for further compression have not been fully explored in this work. Future work could explore the possibility of smaller editors, aiming to reduce the computational burden without compromising performance. This could make EditCoT more accessible and practical for a wider range of applications and deployment scenarios.

\bibliography{custom}

\appendix

\section{Pseudocode Description of EditCoT}

\label{apd:pse}

\begin{algorithm*}
\caption{EditCoT Framework}
\label{algorithm:editcot}
\begin{algorithmic}[1]

\State \textbf{Input:} Query $q$, Original Model $M$, Edited Model $M_{\text{edit}}$, Memory of Facts $K$, Maximum Iterations $N$, Mapping of old facts to new facts $F$
\State \textbf{Output:} Final Answer $A_{\text{final}}$

\State $A_{\text{old}} \gets \text{GenerateAnswer}(M, q)$ \Comment{Generate initial answer}
\State $CoT_{\text{old}} \gets \text{GenerateCoT}(M, q, A_{\text{old}})$ \Comment{Generate initial chain-of-thought}

\For{$i = 1$ to $N$}
    \For{each step $s$ in $CoT_{\text{old}}$}
        \State $F_{\text{new}} \gets \text{RetrieveFacts}(s, K)$ \Comment{Retrieve new fact from memory}
        
        \If{$F$ exists}
            \State $F_{\text{new}} \gets F[F_{\text{new}}]$ \Comment{Map old fact to new fact if applicable}
        \EndIf
        
        \If{$F_{\text{new}} \neq \varnothing$}
            \State \textbf{break} 
        \EndIf
    \EndFor

    \State $is\_Conflict \gets \text{Verify}(M, q, CoT_{\text{old}}, F_{\text{new}})$ \Comment{Verify conflict between CoT and new fact}

    \If{not $is\_Conflict$}
        \State \textbf{break}
    \EndIf

    \State $CoT_{\text{new}} \gets \text{EditChain}(M_{\text{edit}}, q, CoT_{\text{old}}, F_{\text{new}})$ \Comment{Edit chain-of-thought with new fact}
    \State $CoT_{\text{old}} \gets CoT_{\text{new}}$ \Comment{Update old CoT with new one}
\EndFor

\State $A_{\text{final}} \gets \text{GenerateAnswer}(M, CoT_{\text{new}})$ \Comment{Generate final answer based on edited CoT}

\State \Return $A_{\text{final}}$

\end{algorithmic}
\end{algorithm*}
The pseudocode description of EditCoT is shown in Algorithm \autoref{algorithm:editcot}.

\section{Detailed Comparison with the Verify-and-Edit Framework}
\label{cha:vae}
~\citet{zhao2023verifyandeditknowledgeenhancedchainofthoughtframework} introduce a Verify-and-Edit framework that, on the surface, appears similar to our approach as it also involves editing the generated CoT. However, there are several key distinctions: (1) their verification process relies on the self-consistency, which works well for general QA tasks but is less applicable to knowledge editing since the model often shows high confidence in outdated information, leading to high consistency even when incorrect. We conducted tests on MQuAKE-CF-3k, focusing only on the triggering of editing and ignoring subsequent edit results. Using the self-consistency, edits were triggered in only 37\% of cases on the knowledge editing dataset, while our method triggered edits in 95\% of cases.; (2) their editing relies on the model's ability to learn from context, whereas in \autoref{cha:ic}, we demonstrate that our proposed CoT editor significantly outperforms this self-editing approach in knowledge editing tasks; and (3) their method actually resembles Mello's~\cite{zhong2024mquakeassessingknowledgeediting} strategy of decomposing tasks, whereas our approach modifies the entire CoT in each round of editing, allowing for more comprehensive adjustments.

\section{Further Discussion on Relationship Between EditCoT and RAG}

\label{cha:rag}

This section provides a focused discussion between EditCoT and traditional RAG-based approaches, in response to the common concern that EditCoT may overlap significantly with existing retrieval-based methods.

\subsection{Beyond Knowledge-Item Conflicts: Reasoning-Path Level Editing}

\label{apd:cad}

The RAG method occasionally faces conflicts between the knowledge retrieved and the knowledge embedded within the LLM parameters. Techniques like contrastive or context-aware decoding~\cite{li-etal-2023-contrastive, shi-etal-2024-trusting} aim to enforce attention to the retrieved context. However, these approaches generally assume that the LLM merely ignores context, but overlook a deeper issue: the model may integrate the context while still generating a reasoning path that is logically inconsistent with it. For instance, we observe that when the context presents edited knowledge, the model’s reasoning path focuses on clarifying and rectifying the knowledge within the context. In this scenario, the model’s error does not stem from overlooking the context, and therefore contrastive decoding does not address the problem.

In contrast, EditCoT is designed to address precisely these reasoning-path conflicts. Even when the model attends to updated knowledge, its internal chain of thought (CoT) may still follow an outdated or incompatible reasoning structure. EditCoT proposes an iterative editing paradigm that identifies such inconsistencies and revises the CoT accordingly, without requiring internal model access. This makes EditCoT applicable to both open- and closed-source models, such as GPT-4o.

We highlight several key differences between EditCoT and RAG-based contrastive decoding methods:

\begin{itemize}[leftmargin=*, itemsep=0pt]
    \item \textbf{Conflict level:} EditCoT edits reasoning paths iteratively, handling subtle logical inconsistencies, while contrastive decoding typically handles direct knowledge-level conflicts through token-level or intermediate-layer hard enforcement. EditCoT can be seen as a soft method.
    \item \textbf{Model access:} EditCoT does not require internal model access, making it universally applicable even to closed-source large models like GPT-4o (see our experiments on GPT-4o, ~\autoref{tab:gpt4oresult}), whereas contrastive decoding approaches often need white-box access, limiting their broader usability.
    \item \textbf{Pipeline structure:} EditCoT includes a full ICE (In-Context Editing) pipeline—retrieval, conflict detection, and CoT editing—tailored specifically for knowledge editing.  Traditional RAG approaches (including contrastive decoding methods) typically lack such integrated frameworks.
\end{itemize}

\subsection{In-Context Editing and RAG}

Current In-Context Editing (ICE) methods are typically grounded in a retrieval process, and thus can be considered variants of Retrieval-Augmented Generation (RAG). However, general RAG methods are initially developed for open-domain question answering and general knowledge-intensive tasks, and ICE has emerged as a specialized framework tailored to the demands of knowledge editing. These tasks often involve small-scale, precise, and repetitive updates to the model’s underlying knowledge, which are not well-served by standard RAG pipelines.

Unlike RAG, which retrieves relevant information to supplement the model’s parametric knowledge, ICE methods aim to directly guide the model in revising  its reasoning in light of new or corrected knowledge. For example, some ICE variants reformulate edits as the answers to intermediate sub-questions (e.g., PokeMQA~\cite{gu2024pokemqaprogrammableknowledgeediting}), while others construct structured representations for targeted editing (e.g., RAE ~\cite{shi2024retrievalenhancedknowledgeeditinglanguage}). Compared to the standard RAG approach, these adjustments make the ICE method more suitable for knowledge editing scenarios. Building on the limitations identified in these methods, EditCoT further investigates and refines their underlying principles.

However, prior ICE methods do not discuss the specific differences or potential improvements that ICE offers over traditional RAG. To address these limitations, in our main experiments (~\autoref{tab:mainresult}), we extended our baseline comparisons to include not just “naïve RAG” but also two more advanced RAG methods, showing that applying conventional RAG to knowledge editing remains suboptimal without additional adjustments. We explore the reasons for this outcome in \autoref{cha:mainrag}.

\section{Details of Benchmarks}

\autoref{tab:benchmark} shows the statistics of the three benchmarks, including the total number of edits and queries.

\label{apd:benchmark}

\begin{table}[h]
\centering
\caption{Statistics on the number of edits and questions for the benchmarks}
  \resizebox{\linewidth}{!} {
\begin{tabular}{llll}
\toprule
\textbf{Dataset}               & \textbf{Subset}                   & \textbf{Edits} & \textbf{Queries} \\ \hline
\textbf{MQuAKE-CF-3k}          & -                                 & 2785           & 3,000            \\ \hline
\multirow{3}{*}{\textbf{DUNE}} & Scientific Reasoning              & 223            & 1508             \\
                               & Arithmetic Reasoning              & 184            & 1065             \\
                               & New Information                   & 200            & 1000             \\
                               \hline
\textbf{LeKUBE}                & MCQ of the Legal Scenario         & 180            & 180              \\
\textbf{}                      & T/F Questions of Change in Statute & 180            & 642              \\ \bottomrule
\end{tabular}
}

\label{tab:benchmark}
\end{table}

\section{Details of Baselines}
\label{apd:baselines}
We focus on In-Context Editing methods and also compare EditCoT with several traditional model editing techniques:

\textbf{Fine-tuning (FT)}: A full-parameter fine-tuning approach. Since we evaluate Instruct Models, we train the model on question-answer pairs constructed from each dataset's editing instances.

\textbf{Naive RAG}~\cite{lewis2021retrievalaugmentedgenerationknowledgeintensivenlp}: Given an external knowledge base and a retriever, the Retrieval-Augmented Generation (RAG) framework injects relevant knowledge into the model by concatenating retrieved documents into the context, effectively augmenting the language model’s ability to reason and reduce hallucinations~\cite{su-etal-2024-unsupervised,chen2024automatic,su2024mitigating,su2023legalaid}. Following the settings from the original LeKUBE and DUNE papers, we use BM25\cite{robertson2009probabilistic} as the retriever, a retrieval method based on lexical matching, and concatenate 3 and 1 retrieved documents, respectively. For the evaluation of MQuAKE-CF-3k, since the original paper does not test RAG, we still use BM25 as the retriever, setting it to concatenate 5 retrieved documents.

\textbf{RAT}\cite{wang2024ratretrievalaugmentedthoughts}: A RAG method that refines the model-generated reasoning steps through multiple rounds of query rewriting and retrieval. We utilize the official prompt template and ultimately guided the model to derive the final answer based on the original question and the final reasoning steps. All other settings are identical to those in Naive RAG.

\textbf{FLARE}\cite{jiang2023flare}: A dynamic RAG that determines whether to perform retrieval by assessing the uncertainty of the generated sentences during the generation process. We set the hallucination detection threshold at 0.12. All other settings remain the same as in Naive RAG.

\textbf{KN}\cite{dai2021knowledge}: This method updates knowledge by identifying and editing "knowledge neurons" in the model. In our experiments, we set the prompt number \(n\) for neuron identification to 10, the knowledge attribution threshold to 0.2, and the probability of retaining shared neurons to 0.4.

 \textbf{ROME}\cite{meng2023locatingeditingfactualassociations}: A Rank-One Model Editing method that directly modifies key-value pairs in the FFN layers. For our experiments, we uniformly modify the fifth layer and calculate the loss at the final layer of the model. The weight decay is set to \(1 \times 10^{-3}\).

Both KN and ROME methods are implemented using the EasyEdit\cite{wang2023easyedit} library, which integrates several model editing techniques for ease of use. It is important to note that both methods rely on knowledge triples for editing. In the \textit{Scientific Reasoning} and \textit{New Information} subsets of DUNE, the edits consist of longer natural language sequences. We use GPT-4o to extract the relevant knowledge triples.

\textbf{Mello}\cite{zhong2024mquakeassessingknowledgeediting}: Mello performs in-context editing by decomposing the problem into sub-questions and utilizing retrieval. We follow the official setting with a maximum of 4 retrieval rounds and adapt the prompts to Instruct Models. In English datasets, we use Contriever\cite{izacard2022unsuperviseddenseinformationretrieval} as the retriever, following the settings from the original paper. For the Chinese LeKUBE dataset, we translate the prompts into Chinese and use BM25 as the retriever.

\textbf{PokeMQA}\cite{gu2024pokemqaprogrammableknowledgeediting}: Built on top of Mello, PokeMQA adds entity extraction and scope determination to refine the question understanding. We follow the official setting with a maximum of 5 rounds and use their pre-trained Scope-Detector. For DUNE and LeKUBE, where no pre-extracted entities are available, we instruct the model to extract entities in a similar format. Since the official Scope-Detector only supports English, we translate the input sentences for LeKUBE tasks into English before applying the detector.

During the GPT-4o evaluations, we observed difficulties in strictly following the prompt format. To mitigate this issue, we add the system prompt: "Follow the examples below."

Additionally, for Mello, PokeMQA and EditCoT, since LeKUBE evaluates changes in legal provisions, and other baselines can utilize both old and new laws, we concatenate the pre- and post-update legal texts and provide them to the model after retrieval for fair comparison.

\textbf{RAE}\cite{shi2024retrievalenhancedknowledgeeditinglanguage}: RAE constructs knowledge graphs for retrieval and leverages the model to assist in retrieving and pruning the graphs. However, for DUNE and LeKUBE, knowledge graphs similar to those in Wikipedia cannot be constructed, making RAE less generalizable to these tasks.

Additionally, the LLMs we used are implemented by Huggingface Transformers library \cite{wolf-etal-2020-transformers}.

\section{Details of Efficiency Experiments}
\label{apd:efficiency}
All efficiency experiments are conducted on a single NVIDIA A100 (40G) GPU.

RAE’s official implementation uses Wikipedia API calls; hence, we exclude network latency from its reported inference time (including this would place RAE around 46 seconds on the x-axis). Despite this, RAE has a notably longer inference time due to multiple forward propagations for knowledge graph retrieval and pruning, particularly when selecting the next relation, which requires feeding all candidates into the model respectively.

The efficiency of EditCoT is attributed to fewer required iterations: while PokeMQA averages 4.33 iterations, EditCoT needs only 3.65, indicating fewer long-text generation steps. Here a full retrieval-generation cycle is defined as one iteration.

\section{Performance with different max iterations}
\begin{figure}[t]
    \centering
    \includegraphics[width=0.9\linewidth]{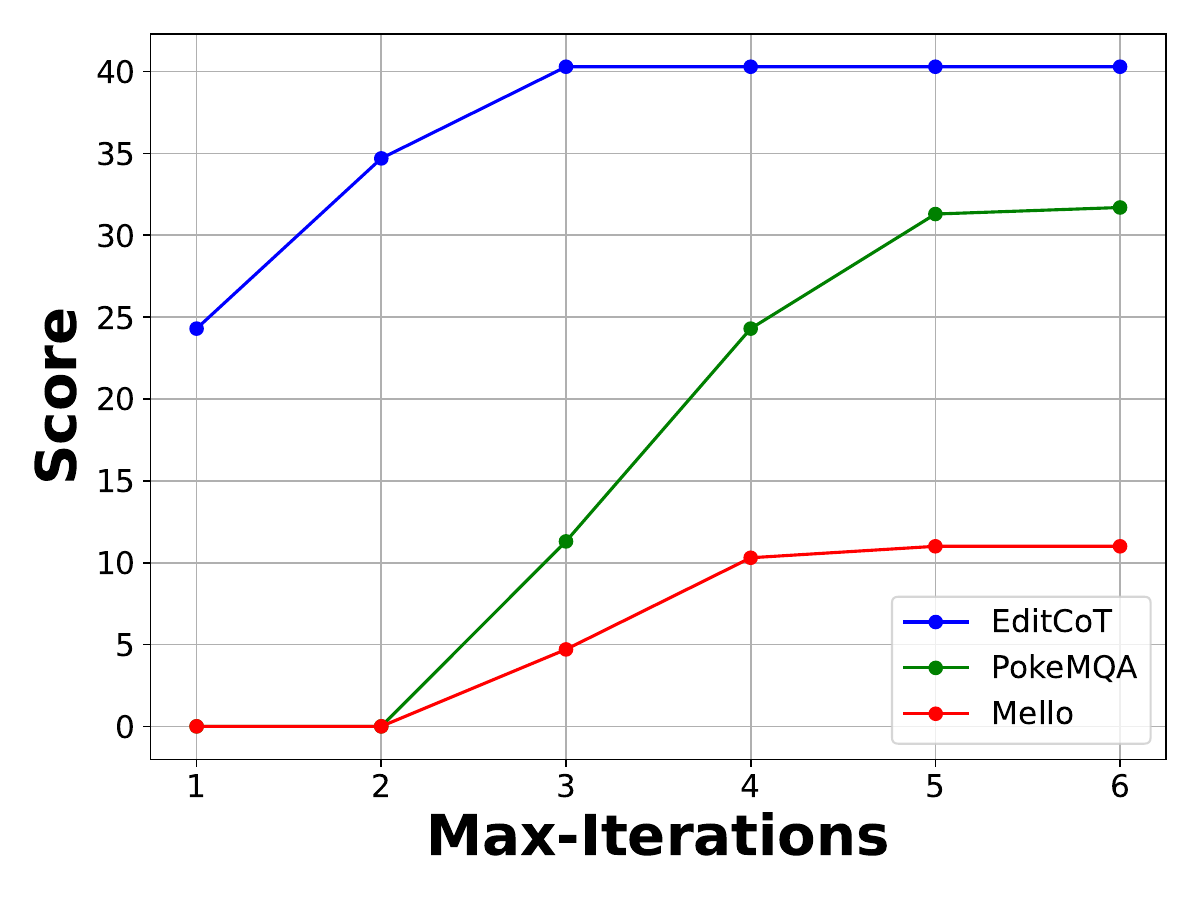}
    \caption{Performance with different max iterations. A full retrieval-generation cycle is defined as one iteration.}
    \label{fig:iterss}
\end{figure}
\label{apd:maxiter}

We explore how varying the maximum number of iterations affects their performance. \autoref{fig:iterss} shows the results. PokeMQA and Mello require more iterations to reach optimal performance, with relatively low accuracy in early iterations. On the other hand, EditCoT outputs a complete chain-of-thought in each iteration, achieving decent performance after the first iteration.

\section{Investigating a SFT Baseline with HotpotQA Knowledge}
\label{app:sft_baseline_hotpotqa_investigation}

To ensure the fairness of our comparison and confirm that any improvements in EditCoT are not due to knowledge leakage from the CoT editor's training process, we employ Supervised Fine-Tuning (SFT) on the Llama3-8B-Instruct, using it to test two baselines. The fine-tuning dataset is constructed using the correct reasoning paths sourced from the CoT Editor’s training dataset, which is derived from HotpotQA. These reasoning paths are formatted into a question-answering (QA) structure suitable for SFT. The fine-tuned model (hereinafter SFT-Llama3-8B-Instruct) possesses a level of knowledge on the HotpotQA domain comparable to that accessible by EditCoT’s CoT Editor.
Subsequently, both the SFT-Llama3-8B-Instruct model and the original Llama3-8B-Instruct model are evaluated on the \textbf{MQuAKE-CF-3k} benchmark. The comparative results are presented in ~\autoref{tab:sft_hotpotqa_investigation_results}.

\begin{table}[h!]
\centering
\caption{Performance comparison on MQuAKE-CF-3k between the original Llama3-8B-Instruct model and its counterpart fine-tuned (SFT) with HotpotQA reasoning paths.}
\resizebox{\linewidth}{!}{
\begin{tabular}{lcc}
\toprule
\textbf{Model Configuration} & \textbf{Mello} & \textbf{PokeMQA} \\
\midrule
SFT Llama3-8B-Instruct & 12.6 & 24.8 \\
Original Llama3-8B-Instruct & 10.0 & 26.1 \\
\bottomrule
\end{tabular}
}
\label{tab:sft_hotpotqa_investigation_results}
\end{table}

We observe that the variations among the inference models lead to only minor differences, which do not translate into a significant advantage or leakage for the newly trained model, further supporting the fairness of our overall design.

\section{Case Studies}
\label{apd:cs}
\begin{figure*}[!t]
    \centering
    \includegraphics[width=\linewidth]{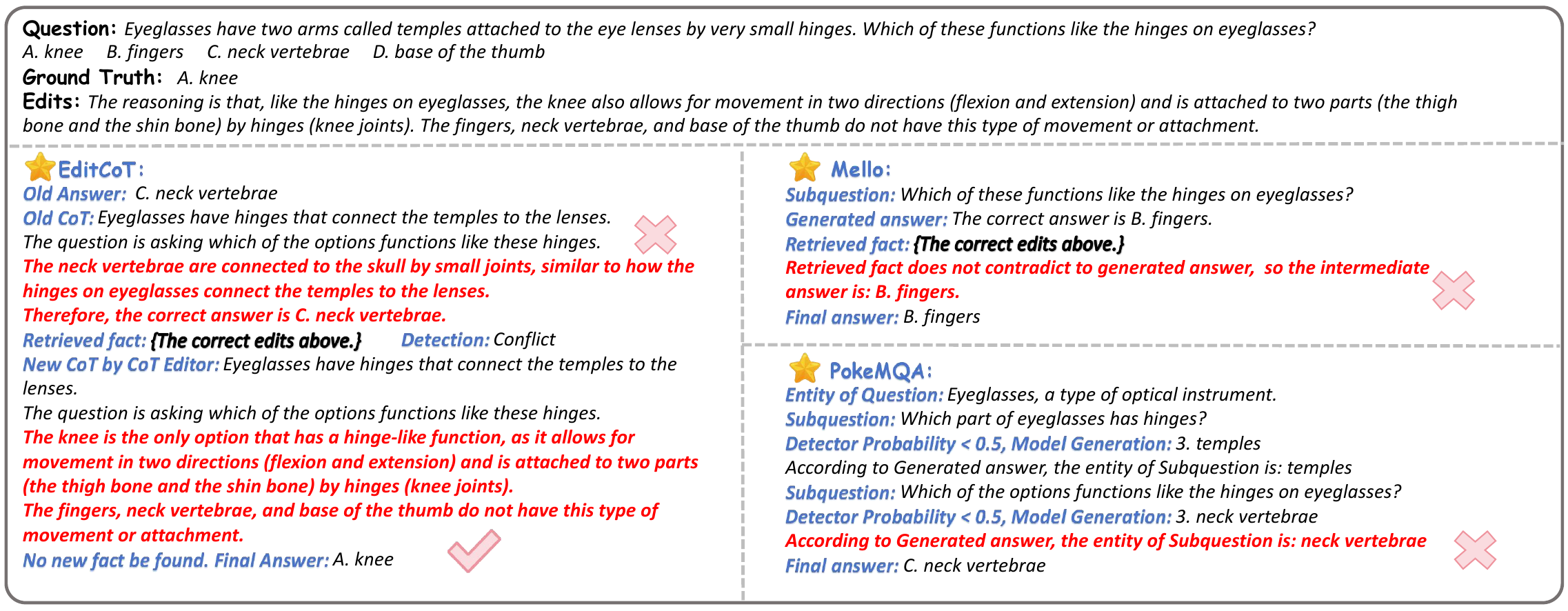}
    \includegraphics[width=\linewidth]{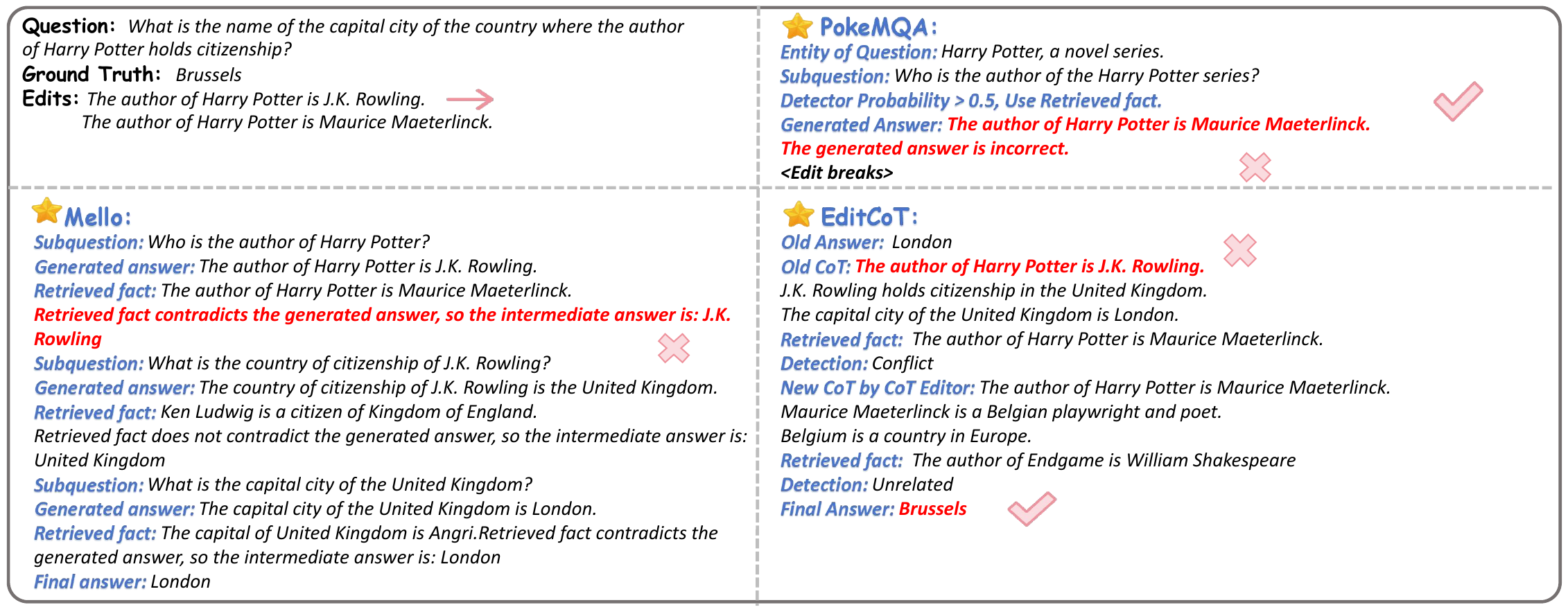}
    \caption{A face-to-face comparison of our proposed EditCoT method with other baseline approaches. The first example comes from the Scientific Reasoning subset of DUNE. The second example comes from MQuAKE-CF-3k. The model here is Meta-Llama-3-8B-Instruct.}
    \label{fig:cs}
\end{figure*}
In \autoref{fig:cs}, we compare EditCoT with Mello and PokeMQA on two questions. The first question from the Scientific Reasoning subset of DUNE shows that EditCoT initially generates an incorrect answer and CoT. However, after retrieval and conflict detection, it modifies the CoT to arrive at the correct solution. In contrast, Mello retrieves a relevant edit but fails to detect any conflict with the initial answer, missing the opportunity for correction. This demonstrates the strength of EditCoT's conflict detection mechanism, which evaluates the entire reasoning chain against retrieved edits. PokeMQA struggles with both sub-questions, producing a repetitive decomposition and failing to trigger an edit.

The second question from MQuAKE-CF-3k again illustrates EditCoT's ability to transition from an incorrect inference to the correct answer through CoT editing. Although both PokeMQA and Mello successfully retrieve and detect the relevant edit (changing the author of Harry Potter to Maurice Maeterlinck), they do not apply it. This indicates that models designed with safety in mind may resist incorporating external knowledge in favor of their internal knowledge. EditCoT addresses this issue by decoupling conflict detection from the editing process and training a model specifically for CoT editing, leading to more effective reasoning updates.

Additionally, \autoref{tab:fail1} and \autoref{tab:fail2} illustrate two failure cases of EditCoT, offering a more comprehensive perspective on its functionality. The error in \autoref{tab:fail1} arises from issues in both the retriever and the conflict detection module. Specifically, the editing process should have correctly terminated after the first edit (i.e., "CoT-2 by CoT Editor"). However, due to suboptimal retriever performance, unrelated examples were retrieved, and conflict detection mistakenly flagged a conflict, leading EditCoT to perform an unnecessary additional edit, which resulted in an incorrect outcome. To maintain fairness in experimental comparisons, EditCoT shares the same retriever with other ICE methods (e.g., Mello), and thus this module was not specifically optimized. It is worth noting that such errors are rare (approximately 1-2 cases out of 200 manually inspected examples), as it is uncommon for both the retriever and conflict detection modules to fail simultaneously. However, this suggests potential future improvements for conflict detection.

In contrast, the error in \autoref{tab:fail2} is entirely due to the original LLM's mistake. The question concerns the origins of certain sports, but the original CoT generated by the model instead provides information about club locations. As a result, the retrieval process produces no results, preventing EditCoT from proceeding as intended. It is important to emphasize that EditCoT is designed to enable reasoning with updated knowledge after editing, rather than to enhance the original model's reasoning capabilities or overall performance. From our manual inspection, such errors are more prevalent and should be attributed to the reasoning limitations of the original model rather than any shortcomings of the CoT Editor itself.

\section{Prompt Templates}

\begin{table*}
\centering
\caption{A failure case example for EditCoT, derived from the results of Llama-3-8B-Instruct on MQuAKE-CF-3k. We highlight the first occurrence of the error in red. The Ground Truth Reasoning Path is the reasoning path provided by the dataset that the edited model is expected to output.}
\resizebox{\textwidth}{!}{
\begin{tabular}{ll}
\hline
\multicolumn{2}{l}{\textbf{Question}: What sport is associated with the employer of John Farrell?} \\ \hline
\multicolumn{2}{l}{\begin{tabular}[c]{@{}l@{}}\textbf{Related Edits:} John Farrell is employed by \uline{BOSTON RED SOX} → \uline{FC Groningen}\\ \\ \textbf{Ground Truth Reasoning Path: }John Farrell is employed by FC Groningen\\ FC Groningen is associated with the sport of association football\end{tabular}} \\ \hline
\textbf{CoT-1 (The original CoT)} &
  \begin{tabular}[c]{@{}l@{}}John Farrell is the manager of the Boston Red Sox.\\ The Boston Red Sox is a team in the sport of Baseball.\end{tabular} \\ \hline
\textbf{Retrieval} &
  \begin{tabular}[c]{@{}l@{}}\textbf{Query:} John Farrell is the manager of the Boston Red Sox.\\ \textbf{Result:} John Farrell is employed by FC Groningen\\ \textbf{Detection:} Conflict\end{tabular} \\ \hline
\textbf{CoT-2 by CoT Editor} &
  \begin{tabular}[c]{@{}l@{}}John Farrell is the manager of the FC Groningen.\\ FC Groningen is associated with the sport of association football.\end{tabular} \\ \hline
\textbf{Retrieval} &
  \begin{tabular}[c]{@{}l@{}}\textbf{Query:} FC Groningen is associated with the sport of association football.\\ \textbf{Result:} \textcolor{red}{AGOVV Apeldoorn is associated with the sport of baseball}\\ \textbf{Detection:} \textcolor{red}{Conflict}\end{tabular} \\ \hline
\textbf{CoT-3 by CoT Editor} &
  \begin{tabular}[c]{@{}l@{}}John Farrell is the manager of AGOVV Apeldoorn.\\ The AGOVV Apeldoorn is a team in the sport of baseball.\end{tabular} \\ \hline
\end{tabular}
}
\label{tab:fail1}
\end{table*}

\begin{table*}
\centering
\caption{A failure case example for EditCoT, derived from the results of Llama-3-8B-Instruct on MQuAKE-CF-3k. We highlight the first occurrence of the error in red. The Ground Truth Reasoning Path is the reasoning path provided by the dataset that the edited model is expected to output.}
\resizebox{\textwidth}{!}{
\begin{tabular}{ll}
\hline
\multicolumn{2}{l}{\textbf{Question:} Which country or location is the origin of the sport associated with NK Nafta Lendava?} \\ \hline
\multicolumn{2}{l}{\begin{tabular}[c]{@{}l@{}}\textbf{Related Edits:} association football was created in \uline{England} → \uline{Hong Kong}\\ \\ \textbf{Ground Truth Reasoning Path:} NK Nafta Lendava is associated with the sport of association football\\ association football was created in Hong Kong\end{tabular}} \\ \hline
\textbf{CoT-1 (The original CoT)} &
  \begin{tabular}[c]{@{}l@{}}NK Nafta Lendava is a football club.\\ The sport associated with NK Nafta Lendava is football (soccer).\\ \textcolor{red}{The country where NK Nafta Lendava is located is Slovenia.}\end{tabular} \\ \hline
\textbf{Retrieval} &
  No results in retrieval. \\ \hline
\textbf{CoT-2 by CoT Editor} &
  \begin{tabular}[c]{@{}l@{}}NK Nafta Lendava is a football club.\\ The sport associated with NK Nafta Lendava is football (soccer).\\ The country where NK Nafta Lendava is located is Slovenia.\end{tabular} \\ \hline
\end{tabular}
}
\label{tab:fail2}
\end{table*}

Since the models we are working with are all Instruct Models, the following prompt template uses "system", "user", and "assistant" to represent different roles.

\subsection{Prompt Templates of Dataset Construction}

\label{apd:prompt1}

First, we use RAG to generate CoT and the answer. The prompt template is as follows:

\begin{tcolorbox}[breakable,colback=lightgray!20,colframe=darkgray!80,title=Dataset Construction - RAG Prompt]
\textbf{User:} 
\{\textit{Top-5 relevant paragraphs from Wikipedia}\}
\\ \hspace*{\fill} \\
Instruction:
You have access to background information that may assist in answering the question. Focus on reasoning through the problem step by step, keeping the explanation concise. Use the token [STEP] to start each thought step and the token [ANSWER] to indicate the final answer. Avoid mentioning or referencing the background information directly in your reasoning.
\\ \hspace*{\fill} \\
Template:

[STEP] ...

[STEP] ...

...

[ANSWER] ...
\\ \hspace*{\fill} \\
Question: \{\textit{A question from HotpotQA}\}
\end{tcolorbox}

Second, we conduct prefix-guided CoT generation. The prompt template is as follows:

\begin{tcolorbox}[breakable,colback=lightgray!20,colframe=darkgray!80,title=Dataset Construction - Prompt of prefix-guided CoT generation]
\textbf{User:} Please answer the following question using a chain-of-thought. Use the token [STEP] to start each thought step, and the token [ANSWER] to indicate the final answer. Keep each step brief and to the point and keep the final answer concise.
\\ \hspace*{\fill} \\
Template:

[STEP] ...

[STEP] ...

...

[ANSWER] ...
\\ \hspace*{\fill} \\
Question: \{\textit{A question from HotpotQA}\}

\textbf{Assistant:} [STEP] \{\textit{The first (k-1) steps of the CoT from RAG}\}

\{\textit{...}\}

\end{tcolorbox}

Then, we detect conflict in the final answers:

\begin{tcolorbox}[colback=lightgray!20,colframe=darkgray!80,title=Dataset Construction - Prompt of Answer Conflict Detection]
\textbf{User:} For the given question Q, and the two provided answers (Answer 1 and Answer 2), determine if the answers are semantically consistent. Choose the most appropriate option from the following:
\\ \hspace*{\fill} \\
A. Yes, they convey similar ideas or information.

B. No, they are different or contradictory.
\\ \hspace*{\fill} \\
Question Q: \{\textit{A question from HotpotQA}\}

Answer 1: \{\textit{The answer generated by RAG}\}

Answer 2: \{\textit{The answer generated by prefix-guided CoT generation}\}
\\ \hspace*{\fill} \\
Your choice:
\end{tcolorbox}

And detect if there is a conflict at the first diverging step between the two CoTs:

\begin{tcolorbox}[colback=lightgray!20,colframe=darkgray!80,title=Dataset Construction - Prompt of CoT Step Conflict Detection]
\textbf{User:} Evaluate the relationship between the following two sentences based on their factual content. Choose the most appropriate option from the following:
\\ \hspace*{\fill} \\
A. The two sentences contain conflicting knowledge.

B. The two sentences support or complement each other.

C. The two sentences are unrelated (no conflict, but no connection).
\\ \hspace*{\fill} \\
Sentence 1: \{\textit{CoT Step k}\}

Sentence 2: \{\textit{CoT Step k*}\}
\\ \hspace*{\fill} \\
Your choice:
\end{tcolorbox}

Finally, we let the model rewrite the key step:

\begin{tcolorbox}[colback=lightgray!20,colframe=darkgray!80,title=Dataset Construction - Rewriting Prompt]
\textbf{User:} Rewrite the following sentence in the style of a Wikipedia data, using formal and objective language. Only express the factual knowledge present in the sentence, without adding any extra information or inventing details. Your output must be one or more sentences, only needs to include direct results without extra words.
\\ \hspace*{\fill} \\
Input:
\\ \hspace*{\fill} \\
Sentence: \{\textit{A CoT step that needs to rewrite}\}
\\ \hspace*{\fill} \\
Output:
\end{tcolorbox}

\subsection{Prompt Templates of EditCoT}

\label{apd:prompt2}

The prompt template for CoT editor is as follows. Note that the template is the same during both the training and inference phases.

\begin{tcolorbox}[colback=lightgray!20,colframe=darkgray!80,title=EditCoT - Prompt for CoT editor]
\textbf{System:} You can edit the following chains of thought based on the new knowledge.

\textbf{User:} Question: \{\textit{A user question}\}

Old Thoughts: \{\textit{A old CoT}\}

New knowledge: \{\textit{A sentence containing new knowledge}\}

Please give me the new chain-of-thought based on the new knowledge.

\textbf{Assistant:} New Thoughts:
\end{tcolorbox}

The prompt templates below are for the original LLM. First, we instruct it to generate an answer directly:

\begin{tcolorbox}[breakable, colback=lightgray!20,colframe=darkgray!80,title=EditCoT - Prompt of Answering directly]

\textbf{User:} Question: What is the capital of the country where Plainfield Town Hall is located?

\textbf{Assistant:} Answer: Washington, D.C.
\\ \hspace*{\fill} \\
\textbf{User:} Question: In which country is the company that created Nissan 200SX located?

\textbf{Assistant:} Answer: Japan
\\ \hspace*{\fill} \\
\textbf{User:} Question: Which continent is the country where the director of "My House Husband: Ikaw Na!" was educated located in?

\textbf{Assistant:} Answer: Asia
\\ \hspace*{\fill} \\
\textbf{User:} Question: Who is the spouse of the US president?

\textbf{Assistant:} Answer: Jill Biden
\\ \hspace*{\fill} \\
\textbf{User:} Question: Who has ownership of the developer of the Chevrolet Corvette (C4)?

\textbf{Assistant:} Answer: General Motors
\\ \hspace*{\fill} \\
\textbf{User:} Question: \{\textit{A user question}\}

\textbf{Assistant:} Answer:
\end{tcolorbox}

The prompt template of get an original CoT from a question and an answer is as follows:

\begin{tcolorbox}[breakable, colback=lightgray!20,colframe=darkgray!80,title=EditCoT - Prompt of Getting the original CoT]

\textbf{User:} Question: What is the capital of the country where Plainfield Town Hall is located?

Answer: Washington, D.C.

Please give the chain of thought based on the question and answer pairs above.

\textbf{Assistant:} Thoughts: Plainfield Town Hall is located in the country of the United States of America.

The capital of United States is Washington, D.C.
\\ \hspace*{\fill} \\
\textbf{User:} Question: In which country is the company that created Nissan 200SX located?

Answer: Japan

Please give the chain of thought based on the question and answer pairs above.

\textbf{Assistant:} Thoughts: Nissan 200SX was created by Nissan.

Nissan is located in the country of Japan.
\\ \hspace*{\fill} \\
\textbf{User:} Question: Which continent is the country where the director of "My House Husband: Ikaw Na!" was educated located in?

Answer: Asia

Please give the chain of thought based on the question and answer pairs above.

\textbf{Assistant:} Thoughts: The director of "My House Husband: Ikaw Na!" is Jose Javier Reyes.

Jose Javier Reyes was educated at De La Salle University.

De La Salle University is located in the country of Philippines.

Philippines is located in the continent of Asia.
\\ \hspace*{\fill} \\
\textbf{User:} Question: Who is the spouse of the US president?

Answer: Jill Biden

Please give the chain of thought based on the question and answer pairs above.

\textbf{Assistant:} Thoughts: The US president is Joe Biden.

The spouse of Joe Biden is Jill Biden.
\\ \hspace*{\fill} \\
\textbf{User:} Question: Who has ownership of the developer of the Chevrolet Corvette (C4)?

Answer: General Motors

Please give the chain of thought based on the question and answer pairs above.

\textbf{Assistant:} Thoughts: The developer of Chevrolet Corvette (C4) is Chevrolet.

Chevrolet is owned by General Motors.
\\ \hspace*{\fill} \\
\textbf{User:} Question: \{\textit{A user question}\}

Answer: \{\textit{An original answer}\}

Please give the chain of thought based on the question and answer pairs above.

\textbf{Assistant:} Thoughts:
\end{tcolorbox}

The prompt template of conflict detection is as follows. We use 5-shot examples and apply a scenario similar to CoT editor.












\begin{tcolorbox}[breakable, colback=lightgray!20,colframe=darkgray!80,title=EditCoT - Prompt of Conflict Detection]
\textbf{System:} You can edit the following chains of thought based on the new knowledge.

\textbf{User:} Question: Which country is the author of \"Moneyball\" a citizen of?

Old Thoughts: The author of Moneyball is Michael Lewis.

Michael Lewis is a citizen of United States of America.

New knowledge: The author of Our Mutual Friend is Charles Dickens

Please give me the new chain-of-thought based on the new knowledge.

\textbf{Assistant:} Relevance of knowledge and chain-of-thought: Unrelated.
\\ \hspace*{\fill} \\
\textbf{User:} Question: What language does the Shadow Home Secretary speak, write, or sign?

Old Thoughts: The Shadow Home Secretary is Diane Abbott.

Diane Abbott speaks the language of English.

New knowledge: The Shadow Home Secretary is Diane Abbott

Please give me the new chain-of-thought based on the new knowledge.

\textbf{Assistant:} Relevance of knowledge and chain-of-thought: Support.
\\ \hspace*{\fill} \\
\textbf{User:} Question: What is the notable work of the author of \"This Side of Paradise\"?

Old Thoughts: The author of This Side of Paradise is F. Scott Fitzgerald.

F. Scott Fitzgerald is famous for The Great Gatsby.

New knowledge: The author of This Side of Paradise is Geoffrey Chaucer

Please give me the new chain-of-thought based on the new knowledge.

\textbf{Assistant:} Relevance of knowledge and chain-of-thought: Contradict.

New Thoughts: The author of This Side of Paradise is Geoffrey Chaucer.

Geoffrey Chaucer is famous for Shaman King.
\\ \hspace*{\fill} \\
\textbf{User:} Question: Who is the head of government of the country where Roy Ascott holds citizenship?

Old Thoughts: Roy Ascott is a citizen of United Kingdom.

The name of the current head of the United Kingdom government is Boris Johnson.

New knowledge: Roy Ascott is a citizen of United States of America

Please give me the new chain-of-thought based on the new knowledge.

\textbf{Assistant:} Relevance of knowledge and chain-of-thought: Contradict.

New Thoughts: Roy Ascott is a citizen of United States of America.

The name of the current head of the United States of America government is Ole Johan Vierdal.
\\ \hspace*{\fill} \\
\textbf{User:} Question: Who is the spouse of the performer of the song \"Don't Be Cruel\"?

Old Thoughts: Don't Be Cruel was performed by Elvis Presley.

Elvis Presley is married to Priscilla Presley.

New knowledge: Don't Be Cruel was performed by Elvis Presley

Please give me the new chain-of-thought based on the new knowledge.

\textbf{Assistant:} Relevance of knowledge and chain-of-thought: Support.
\\ \hspace*{\fill} \\
\textbf{User:} Question: \{\textit{A user question}\}

Old Thoughts: \{\textit{An old CoT}\}

New knowledge: \{\textit{A sentence containing new knowledge}\}

Please give me the new chain-of-thought based on the new knowledge.

\textbf{Assistant:} Relevance of knowledge and chain-of-thought:
\end{tcolorbox}

The prompt of getting final answer is as follows. We instruct the original LLM to pay more attention to the new CoT.

\begin{tcolorbox}[breakable, colback=lightgray!20,colframe=darkgray!80,title=EditCoT - Prompt of Getting Final Answer]
\textbf{System:} Suppose the world knowledge has been updated. Answer the question based on the [New Thoughts] rather than your own knowledge. Don't analyze whether the reasoning of [New Thoughts] is correct. Output the most concise answer directly.

\textbf{User:} Question: In which location are the headquarters of the developer of Android Jelly Bean situated?

[New Thoughts]: Android Jelly Bean was developed by Google.

The headquarters of Google is located in the city of Googleplex.

\textbf{Assistant:} Answer from [New Thoughts]: Googleplex
\\ \hspace*{\fill} \\
\textbf{User:} Question: Who is the head of government in the city where Husky Energy's headquarters is located?

[New Thoughts]: The headquarters of Husky Energy is located in the city of Calgary.

The name of the current head of the Calgary government is Jyoti Gondek.

\textbf{Assistant:} Answer from [New Thoughts]: Jyoti Gondek
\\ \hspace*{\fill} \\
\textbf{User:} Question: Who is the head of state of the country of citizenship of Theodoros Angelopoulos?

[New Thoughts]: Theodoros Angelopoulos is a citizen of Greece.

The name of the current head of state in Greece is Aikaterini Sakellaropoulou.

\textbf{Assistant:} Answer from [New Thoughts]: Aikaterini Sakellaropoulou
\\ \hspace*{\fill} \\
\textbf{User:} Question: Who is the head of government in the location where Vulcan Inc. is headquartered?

[New Thoughts]: The headquarters of Vulcan Inc. is located in the city of Seattle.

The name of the current head of the Seattle government is Bruce Harrell.

\textbf{Assistant:} Answer from [New Thoughts]: Bruce Harrell
\\ \hspace*{\fill} \\
\textbf{User:} Question: Who is the head of government of the country where Greg Combet holds citizenship?

[New Thoughts]: Greg Combet is a citizen of Australia.

The name of the current head of the Australia government is Anthony Albanese.

\textbf{Assistant:} Answer from [New Thoughts]: Anthony Albanese
\\ \hspace*{\fill} \\
\textbf{User:} Question: \{\textit{A question}\}

[New Thoughts]: \{\textit{A CoT}\}

\textbf{Assistant:} Answer from [New Thoughts]:

\end{tcolorbox}

\section{Licensing}
Qwen2.5-14B-Instruct is released under the Apache License 2.0. Meta-Llama-3-8B-Instruct is released under the META LLAMA 3 COMMUNITY LICENSE. KN, ROME, FT, FLARE are released under the MIT license. Contriever is released under the CC BY-SA 4.0 License.

The datasets MQuAKE, LeKUBE, and DUNE are released under the MIT license. This paper's research objective is academic exploration, which aligns with the terms of this license.

\end{document}